\documentclass[10pt,twocolumn,letterpaper]{article}

\usepackage{cvpr}              

\usepackage[accsupp]{axessibility}
\usepackage{multirow}
\usepackage[utf8]{inputenc}
\usepackage{booktabs} 
\usepackage[table]{xcolor} 

\definecolor{myPurple}{rgb}{0.4, .0, .8}
\definecolor{myGreen}{rgb}{0, 0.6, .3}
\definecolor{myRed}{rgb}{0.8, .2, .2}
\definecolor{myOrange}{rgb}{0.8, 0.45, 0.0}
\definecolor{myBlue}{rgb}{.0, .0, 1.0}
\definecolor{myBlue2}{rgb}{.0, 1.0, 1.0}
\definecolor{myBlack}{rgb}{.0, .0, 0.0}
\definecolor{darkmidnightblue}{rgb}{0.0, 0.2, 0.4}
\definecolor{MyGreen}{rgb}{0.02,0.5,0.02}

\definecolor{bestcolor}{HTML}{F59194}
\definecolor{secondcolor}{HTML}{FAC791}
\definecolor{thirdcolor}{HTML}{FEF3C7}
\definecolor{white}{HTML}{FFFFFF}
\newcommand{\other}[1]{%
  \multicolumn{1}{>{\columncolor{white}}c}{#1}%
}
\newcommand{\best}[1]{%
  \multicolumn{1}{>{\columncolor{bestcolor}}c}{\textbf{#1}}%
}
\newcommand{\second}[1]{%
  \multicolumn{1}{>{\columncolor{secondcolor}}c}{#1}%
}
\newcommand{\third}[1]{%
  \multicolumn{1}{>{\columncolor{thirdcolor}}c}{#1}%
}

\definecolor{cvprblue}{rgb}{0.21,0.49,0.74}
\usepackage[pagebackref,breaklinks,colorlinks,allcolors=cvprblue]{hyperref}

\def\paperID{5588} 
\def\confName{CVPR}
\def\confYear{2026}

\title{GaussianZoom: Progressive Zoom-in Generative 3D Gaussian Splatting \\ with Geometric and Semantic Guidance
\vspace{-1em}}

\author{Jiale Shi\footnotemark[1] \quad
Jiarui Hu\footnotemark[1] \quad
Zesong Yang \quad
Kaixuan Luan \quad
Hujun Bao \quad
Zhaopeng Cui\footnotemark[2]
\\
 State Key Lab of CAD \& CG, Zhejiang University
 \\
 \vspace{0.25em}
 \small{Project Page: \url{https://zju3dv.github.io/GaussianZoom/}}
}

\begin{document}
\setlength{\abovedisplayskip}{0.4 em}
\setlength{\belowdisplayskip}{0.4 em}
\twocolumn[{
\renewcommand\twocolumn[1][]{#1}
\maketitle
    \vspace{-1em}
    \centering
    \includegraphics[width=\textwidth]{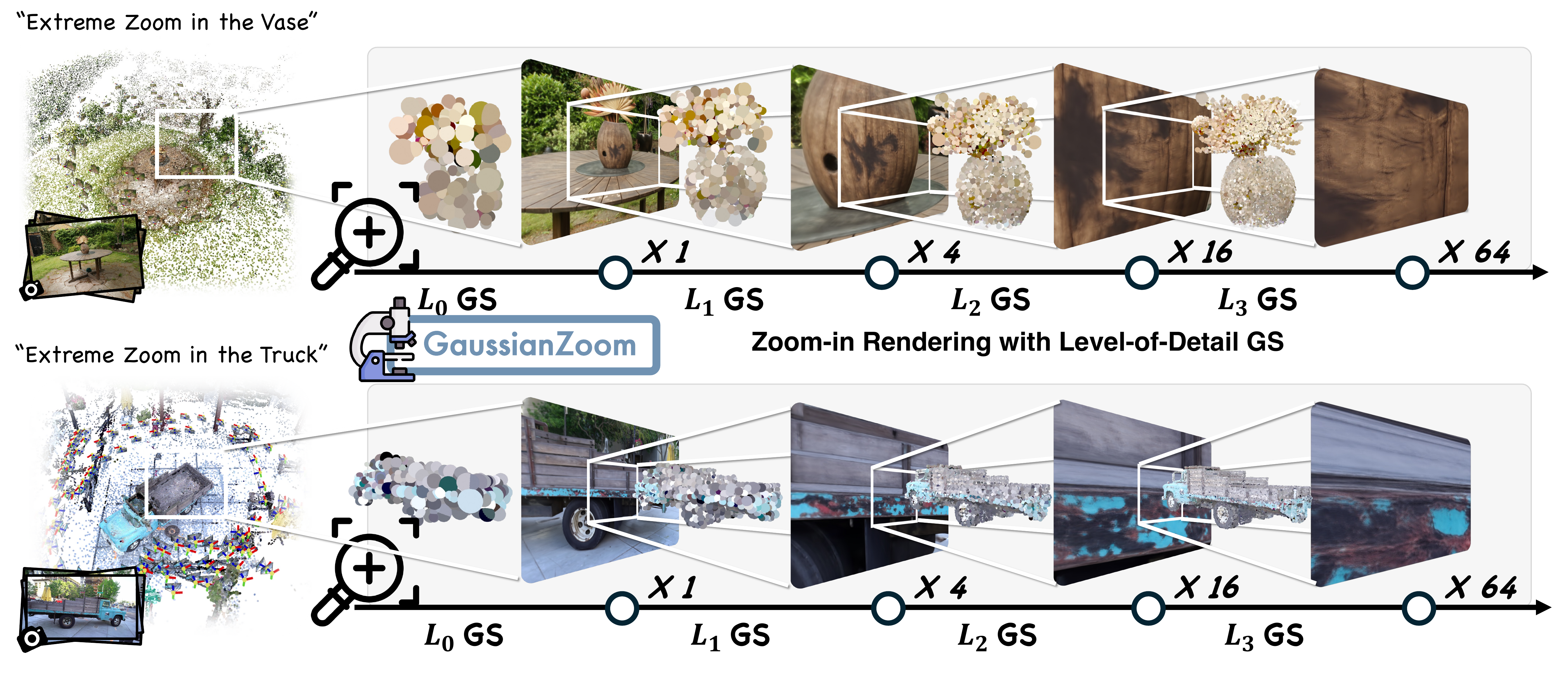}
    \vspace{-1em}
    \captionof{figure}{
    \textbf{GaussianZoom} progressively magnifies 3D scenes from low-resolution inputs, reconstructing them into multi-view consistent and detail-rich representations. The expandable continuous Level-of-Detail hierarchy organizes primitives across scales, enabling smooth and alias-free rendering throughout the zoom-in process. Please refer to the supp. material for more vivid video demonstrations.
    }
    \label{fig:teaser}
    \vspace{1.0 em}
}]

\maketitle
\renewcommand{\thefootnote}{\fnsymbol{footnote}}
\footnotetext[1]{\noindent Equal contribution.}
\footnotetext[2]{\noindent Corresponding author.}
\begin{abstract}
We introduce GaussianZoom, a generative zoom-in 3D reconstruction system with an iterative progressive framework that combines geometry-consistent scene modeling and multi-scale semantic reasoning to enable high-fidelity extreme zoom-in rendering from low-resolution inputs.
To achieve this, we develop a novel multi-view consistent super-resolution module with depth-based feature warping and VLM-driven detail synthesis, ensuring accurate multi-view correspondence while enriching fine-scale appearance beyond the observed resolution.
To support zooming across large magnification ranges, we further introduce a new expandable continuous Level-of-Detail hierarchy that dynamically modulates Gaussian visibility for smooth, alias-free cross-scale rendering. Experiments on Mip-NeRF360 and Tanks\&Temples demonstrate that GaussianZoom achieves superior perceptual quality, multi-view consistency, and robustness under extreme magnification, establishing a strong baseline for generative zoom-in 3D scene reconstruction.
\end{abstract}    

\section{Introduction}
\label{sec:intro}

Reconstructing high-fidelity 3D scenes from images is a fundamental problem in computer vision and graphics, supporting applications such as immersive VR/AR, digital content creation, and embodied perception. While recent advances in 3D Gaussian Splatting (3DGS)~\cite{kerbl20233d} have demonstrated impressive rendering quality and real-time performance, their reconstruction fidelity remains inherently constrained by the resolution and clarity of the input images. When the captured views are low-resolution due to distant viewpoints or hardware constraints, the reconstructed 3D scenes exhibit blurry textures, missing fine structures. These limitations become increasingly pronounced under zoom-in rendering, where users expect coherent geometric details and semantically meaningful textures at progressively higher magnifications.

Traditional 3D super-resolution (SR) attempts to address this issue by employing 2D image or video SR models on input images before 3D reconstruction. However, these approaches inherently lack cross-view geometric consistency, because single-image SR independently sharpens each frame without enforcing geometric alignment~\cite{feng2024srgs,yu2024gaussiansr,lee2024disr,xie2024supergs,chen2025bridging}, 
while flow-based video SR suffers from optical-flow failures under occlusion and large parallax~\cite{shen2024supergaussian,ko2025sequence}.
Moreover, such models are limited to enhancing details already observable in the low-resolution input and cannot generate plausible new details. As a result, existing 3D super-resolution methods often produce inconsistent artifacts and fail to reveal fine-scale semantics beyond the captured resolution.
These limitations suggest that zoom-in 3D reconstruction is fundamentally a progressive generative process rather than a single-shot upsampling problem.
At each zoom step, the system should remain anchored in geometry, i.e., preserving accurate 3D structure and cross-view alignment, 
while simultaneously enriching appearance with semantically plausible details guided by high-level scene understanding. 
In essence, zooming into a 3D scene reflects a continuous transition from reconstruction toward generation.

To this end, we propose \textbf{GaussianZoom}, a progressive zoom-in generative 3D Gaussian Splatting framework that performs iterative coupling between geometry-consistent modeling  and semantic-guided detail synthesis. At each zoom step, a depth-based feature warping module replaces conventional flow-based warping in video SR model with accurate geometric correspondence derived from the coarse, geometry regularized 3DGS reconstruction~\cite{zhang2024rade}, ensuring cross-view consistency. Concurrently, a vision-language model (VLM) infers high-level semantic cues from multi-scale renderings, guiding the SR model to synthesize new, semantically consistent details beyond the observed resolution. The synthesized zoomed-in images then supervise the next-step 3DGS optimization, 
forming a progressive refinement pipeline that incrementally enriches scene geometry and appearance.

Beyond iterative refinement,
we introduce an expandable continuous Level-of-Detail (LoD) representation that elevates LoD from a discrete efficiency-oriented mechanism to a continuous generative scaffold.
In contrast to conventional LoD structures~\cite{kerbl2024hierarchical,ren2024octree,kulhanek2025lodge,seo2024flod} which are primarily designed for static reconstruction and efficient rendering, our LoD hierarchy grows with the zoom-in process, 
connecting coarse-scale reconstruction with finer-scale detail synthesis.
Moreover, rather than performing abrupt scale switching, our method dynamically adjusts the visibility of Gaussian primitives according to their scale projection coefficient, enabling alias-free rendering and smooth transitions across scales.
Each zoom-in step introduces a new LoD layer populated with semantically generated high-frequency details, while previous layers retain coarse appearance and preserve global structure.
This scale-aware generative hierarchy ensures scene-level coherence as the resolution increases, enabling finer layers to progressively inject semantically consistent VLM-guided details.

Overall, our contributions can be summarized as follows:
\begin{itemize}
\item We present \textbf{GaussianZoom}, a generative zoom-in 3D reconstruction system that integrates geometry-consistent scene modeling 
with multi-scale semantic reasoning to enable high-fidelity extreme zoom-in rendering from low-resolution inputs.


\item We introduce a novel multi-view consistent super-resolution module with depth-based feature warping and VLM-driven detail synthesis, ensuring accurate multi-view correspondence while enriching fine-scale appearance beyond the observed resolution.

\item We develop a new expandable continuous Level-of-Detail representation that dynamically modulates Gaussian visibility across scales, enabling alias-free rendering and smooth cross-scale transitions.

\item Experiments across multiple datasets demonstrate that GaussianZoom consistently outperforms prior approaches, delivering superior multi-view consistency, super-resolution quality, and 3DGS photorealism.

\end{itemize}

\section{Related Work}
\label{sec:relatedwork}

\noindent\textbf{2D Super-Resolution.}
Image super-resolution (ISR) has evolved from early CNN-based models such as EDSR~\cite{lim2017enhanced} and RCAN~\cite{zhang2018image}, which optimize pixel-wise fidelity but often yield over-smoothed results, to perceptual and adversarial formulations such as SRGAN~\cite{ledig2017photo}, ESRGAN~\cite{wang2018esrgan}, and Real-ESRGAN~\cite{wang2021real}. More recently, diffusion-based SR approaches have emerged as strong generative priors for texture restoration, including StableSR~\cite{wang2024exploiting}, ResShift~\cite{yue2023resshift}, and OSEDiff~\cite{wu2024one}.
Video SR (VSR) further emphasizes temporal coherence. Classical methods such as EDVR~\cite{wang2019edvr} and BasicVSR++~\cite{chan2022basicvsr++} rely on optical-flow or recurrent propagation but are sensitive to flow errors. Diffusion-based VSR models such as Upscale-A-Video~\cite{zhou2024upscale} and DLoRAL~\cite{sun2025one} improve perceptual detail but often struggle with temporal stability.

\noindent\textbf{3D Super-Resolution.}
Applying 2D SR before 3D reconstruction commonly leads to cross-view inconsistencies. SRGS~\cite{feng2024srgs} sharpens input views but lacks explicit geometric alignment. GaussianSR~\cite{yu2024gaussiansr} and DiSR-NeRF~\cite{lee2024disr} introduce diffusion priors for consistency at the cost of expensive multi-step score distillation. SuperGS~\cite{xie2024supergs} mitigates pseudo-label noise but still relies on per-view enhancement, while SuperGaussian~\cite{shen2024supergaussian} employs video SR yet remains sensitive to large motions and occlusions. Sequence Matters~\cite{ko2025sequence} improves temporal propagation with PSRT~\cite{shi2022rethinking} but continues to inherit flow inaccuracies. 3DSR~\cite{chen2025bridging} integrates diffusion directly into 3D Gaussian Splatting but suffers from high computational overhead due to multiple sampling steps. In contrast, our method leverages reconstructed depth for geometry-guided warping, enabling accurate cross-view correspondence.

\noindent\textbf{Semantic Detail Enhancement.}
Beyond fixed-scale SR, recent works explore text-conditioned zoom-in synthesis for enriching semantic details. Generative Powers of Ten~\cite{wang2024generative} highlights the potential of text-to-image models to produce coherent imagery under extreme magnification, but its single-pass generation makes cross-view consistency difficult to control. Chain-of-Zoom (CoZ)~\cite{kim2025chain} addresses this with a progressive zoom strategy guided by VLM-inferred prompts. Similar to CoZ, we employ a VLM to infer fine-scale semantic cues, but unlike 2D zoom-in frameworks, our approach operates in 3D and provides multi-scale geometric context, enabling semantically and geometrically consistent detail synthesis across views.

\noindent\textbf{Level-of-Detail Gaussian Splatting.}
Recent extensions of 3DGS incorporate Level-of-Detail (LoD) structures to improve rendering efficiency through hierarchical or octree-based Gaussian representations~\cite{kerbl2024hierarchical,ren2024octree,kulhanek2025lodge,seo2024flod}. These methods prioritize computational savings by selecting appropriate subsets of primitives based on camera distance. In contrast, our LoD formulation emphasizes scale-aware coherence: finer primitives are progressively introduced during the zoom-in process while existing ones remain fixed. This design ensures smooth transitions across scales and provides a multi-scale structural prior that supports progressive detail synthesis rather than serving solely as a rendering optimization mechanism.

\section{Preliminaries}

\label{sec:preliminary}
\noindent\textbf{3D Gaussian Splatting.}
The standard 3D Gaussian Splatting (3DGS)~\cite{kerbl20233d} framework represents the scene with a set of explicit 3D Gaussian primitives, each parameterized by a center $\mu$ and a full covariance matrix $\Sigma$:
\begin{equation}
G(x) = e^{-\frac{1}{2}(x-\mu)^{\top}\Sigma^{-1}(x-\mu)}.
\end{equation}
For differentiable rendering optimization, $\Sigma$ is decomposed into a scaling matrix $S$ and rotation matrix $R$, i.e., $\Sigma = R S S^\top R^\top$,
where $S$ and $R$ are parameterized by a 3D scale vector $s$ and a quaternion $q$ respectively.
Given a viewing transformation $W$, the 3D Gaussians are projected to the image plane, and we obtain the 2D covariance matrix $\Sigma'$ and 2D center location $\mu'$ as: 
\begin{equation}
        \Sigma' = JW\Sigma W^\top J^\top, \quad
        \mu' = JW\mu,
\end{equation}
where $J$ is the Jacobian of the affine approximation of the projective
transformation.
Then the rendered color $C$ can be computed by alpha-blending the projected $N$ Gaussians sorted by depths.
\begin{equation}
    C = \sum_{i \in N}T_ic_i\alpha_i,
    \quad
    T_i = {\textstyle \prod_{j=1}^{i-1} (1 - \alpha_j)},
\label{eq:alpha_blending}
\end{equation}
with alpha $\alpha_i$ defined as: 
\begin{equation}
\alpha_i = o_i e^{-\frac{1}{2}(x - \mu'_i)^{\top} \Sigma'^{-1}_i (x - \mu'_i)}.
\end{equation}
Here, $o_i$ and $c_i$ denote the learned opacity and color encoded by spherical harmonics for each primitive.
\begin{figure}[t]
  \centering
  \setlength{\tabcolsep}{1pt} 
  \renewcommand{\arraystretch}{0.5}
  \begin{subfigure}[t]{0.22\textwidth}
    \centering
    \includegraphics[width=\linewidth]{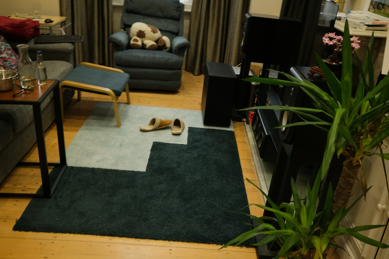}
    \caption{Reference view}
  \end{subfigure}
  \hspace{4pt}
  \begin{subfigure}[t]{0.22\textwidth}
    \centering
    \includegraphics[width=\linewidth]{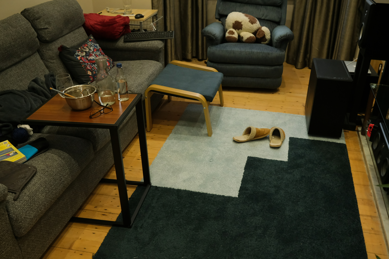}
    \caption{Source view}
  \end{subfigure}

  \begin{subfigure}[t]{0.22\textwidth}
    \centering
    \includegraphics[width=\linewidth]{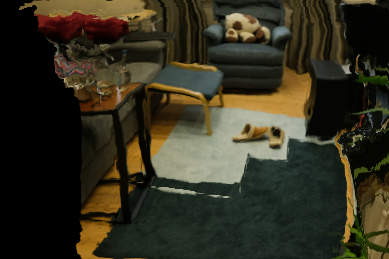}
    \caption{Warped by optical flow}
  \end{subfigure}
  \hspace{4pt}
  \begin{subfigure}[t]{0.22\textwidth}
    \centering
    \includegraphics[width=\linewidth]{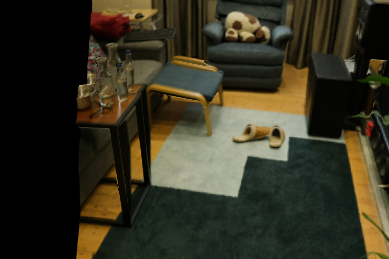}
    \caption{Warped by reconstructed depth}
  \end{subfigure}
  \vspace{-0.8em}
  \caption{
    Comparison between flow-based and depth-based warping. The proposed depth-guided alignment achieves geometrically consistent correspondences across views and effectively suppresses ghosting artifacts.
  }
  \label{fig:flow_warp}
  \vspace{-1em}
\end{figure}
\section{Methods}
\label{sec:methods}
\begin{figure*}[!t] 
    \centering
    \scriptsize
    \vspace{-2em}
    \includegraphics[width=1.0 \textwidth]{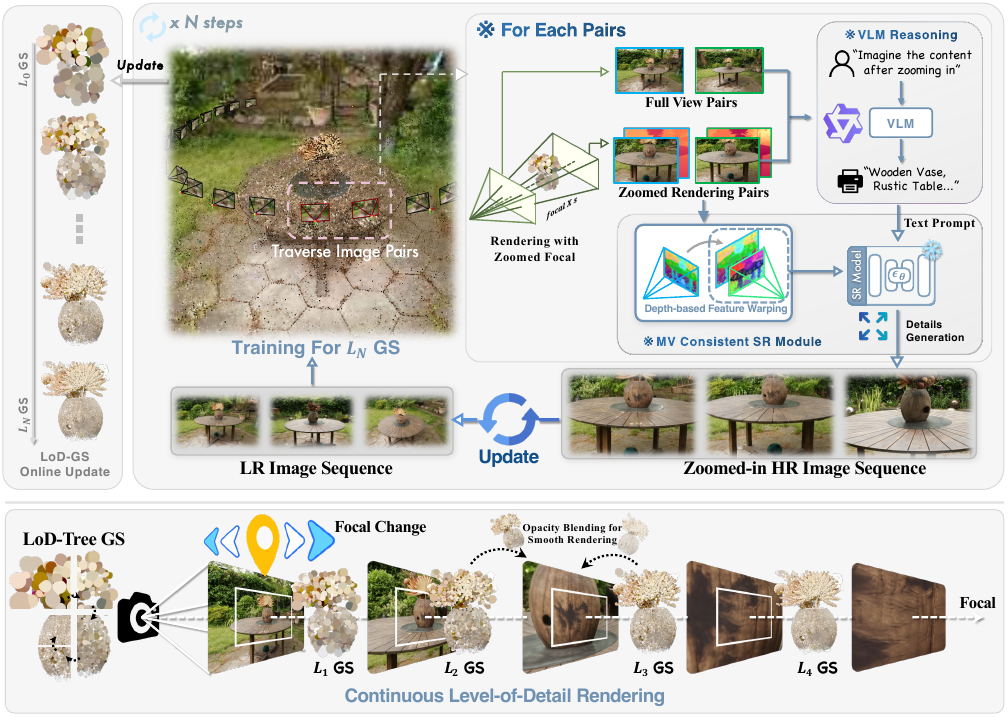}
     \vspace{-2em} 
\caption{\textbf{Method overview.} Our framework jointly leverages geometry-aware alignment, semantic priors, and a continuous Level-of-Detail (LoD) representation to perform generative zoom-in reconstruction. Starting from a coarse 3D Gaussian Splatting model, we derive per-view depth maps that enable depth-based feature warping, providing accurate multi-view correspondence. In parallel, coarse and zoomed-in renderings are processed by a vision-language model to infer semantic cues describing fine-scale appearance. These geometry-aligned features and semantic descriptions together condition the super-resolution network, synthesizing high-resolution zoomed views with plausible, view-consistent details. The resulting images are used to update a continuous LoD hierarchy, where opacity of each primitive is dynamically adjusted to enable alias-free rendering and smooth transitions across zoom levels.}

    \label{fig:pipeline}
    \vspace{-2em}
\end{figure*}

As illustrated in Fig.~\ref{fig:pipeline}, given posed low-resolution image sequences, we progressively reconstruct the scene through a generative zoom-in process. At each zoom step, a unified multi-view consistent super-resolution module (Sec.~\ref{sec:method_sr_module}) combines depth-guided feature warping, derived from the geometry-regularized 3DGS, with vision-language model driven semantic conditioning to synthesize high-resolution views that are both geometrically aligned and semantically enriched beyond the captured resolution. The synthesized zoomed-in images are then used to refine the underlying Gaussian representation at the corresponding scale, while an expandable and continuous Level-of-Detail hierarchy (Sec.~\ref{sec:method_lod}) organizes multi-scale Gaussian primitives and dynamically modulates their visibility, enabling alias-free rendering and smooth transitions across zoom levels. 
\subsection{Multi-View Consistent SR Module}
\label{sec:method_sr_module}
To achieve multi-view consistent and semantically enriched zoom-in reconstruction, we integrate depth-based feature warping with VLM-driven detail synthesis within a unified super-resolution module.

\noindent\textbf{Depth-based Feature Warping.}
We start from flow-based video super-resolution (VSR) frameworks that align neighboring frames through optical flow, typically estimated by pretrained models such as SpyNet~\cite{ranjan2017optical}. Although such flow-guided alignment performs reasonably well under moderate motion and small viewpoint changes, it relies solely on appearance correspondence and thus easily fails in the presence of occlusions, textureless regions, or large parallax. These limitations lead to inaccurate feature alignment and inconsistent generation across views (Fig.~\ref{fig:flow_warp}).

We therefore propose a geometry-aware depth-based warping mechanism that leverages the reconstructed depth from 3D Gaussian Splatting. A geometrically consistent low-resolution Gaussian model $\mathcal{G}$ is first optimized from input LR images ${I_i}$, producing reliable per-view depth maps $\mathbf{D}_i$ that serve as explicit geometric priors. 

Given camera intrinsics $\mathbf{K}_i,\mathbf{K}_j$ and extrinsics $\mathbf{P}_i$, $\mathbf{P}_j$ for frames $i$ and $j$, 
the geometric correspondence between a pixel $\mathbf{p}=(u,v,1)$ in view $j$ and its projection 
$\mathbf{p}'=(u',v',1)$ in view $i$ is expressed as:
\begin{equation}
\mathbf{p}'^{\top} \mathbf{D}'_i 
= 
\mathbf{K}_i\, \mathbf{P}_i \mathbf{P}_j^{-1} \mathbf{K}_j^{-1}
\mathbf{p} \mathbf{D}_j ,
\end{equation}
where $\mathbf{D}'_i$ denotes the reprojected depth in the coordinate frame of camera $i$.
This defines a dense geometric warp $W_{j \rightarrow i}$, which is applied to feature maps:
\begin{equation}
\tilde{\mathbf{F}}_{i} = W_{j \rightarrow i}(\mathbf{F}_{j}).
\end{equation}

By anchoring alignment to reconstructed geometry, this depth-guided warping ensures accurate cross-view correspondence and naturally resolves occlusions and parallax, yielding stable and consistent feature propagation across views.

\noindent\textbf{VLM-Driven Detail Synthesis.}
Although depth-based feature warping improves multi-view consistency, it remains fundamentally constrained by the observable content in LR inputs. To introduce semantically meaningful fine-scale details, we incorporate priors from vision-language model into the super-resolution pipeline.

At each zoom step, we render a coarse-scale view containing global semantics and a zoomed-in view highlighting regions with insufficient high-frequency detail. These paired renderings are processed by a vision-language model, which infers a textual description of fine-scale attributes such as materials and textures.

The text description $c$, together with the multi-view consistent features $\tilde{\mathbf{F}}_{i}$ obtained through depth-guided warping and the original feature representation $\mathbf{F}_i$, provides semantic and geometric conditioning for the super-resolution module to synthesize fine-scale details. We express the super-resolution process as:
\begin{equation}
    I^{\mathrm{sr}}_i =
    \mathcal{S}\!\left(
       \mathbf{F}_i,\; \tilde{\mathbf{F}}_i,\; c
    \right),
    \label{eq:sr}
\end{equation}
where $\mathcal{S}(\cdot)$ denotes the super-resolution network.
The synthesized HR image $I^{\mathrm{sr}}_i$ sharpens visible structures while introducing plausible semantic details consistent with both the global context and local zoomed content. These HR outputs then serve as supervision for updating the Gaussian representation at the corresponding zoom level.

Through this iterative combination of 3D reconstruction, geometric alignment, semantic reasoning, and detail synthesis, the framework progressively enhances visual fidelity while maintaining stable multi-view consistency across zoom levels.

\subsection{Continuous LoD Representation}
\label{sec:method_lod}
We further introduce an expandable and continuous Level-of-Detail (LoD) representation that treats LoD not as a static, efficiency-driven structure, but as a generative scaffold that grows alongside the progressive zoom-in reconstruction. Unlike conventional LoD schemes, which switch among pre-defined levels, our formulation enables each Gaussian primitive to dynamically adjust its opacity based on its scale projection coefficient, thereby supporting alias-free rendering and smooth cross-scale transitions without explicit level switching.

We define the scale projection coefficient as
\begin{equation}
    \psi=\frac{d}{f},
\end{equation}
where $d$ denotes the distance from the camera center to the primitive center, and $f$ is the focal length of the camera. This coefficient reflects how a primitive's world-space extent projects onto the image plane. A smaller $\psi$ indicates that the primitive occupies a larger screen-space footprint and thus should be represented by finer, high-resolution components.

During rendering, we compare the current $\psi'$ (under the rendering camera) with the stored $\psi$ (at the scale where the primitive was created). If $\psi'/\psi$ exceeds the zoom factor $s$, the primitive becomes under-resolved, and finer-level representations are favored to capture the necessary high-frequency detail. Conversely, when $\psi'/\psi$ falls below $1/s$, the primitive sufficiently covers its projected footprint, and its contribution is increased while finer-level components are suppressed.
To achieve smooth transitions, we adjust the primitive's opacity using a logarithmic attenuation function:
\begin{equation}
    w(\psi'/\psi)=\max(0, 1-|\log_s(\psi'/\psi)|),
\end{equation}
where $s$ denotes the scale factor of each zoom step.
This formulation yields a continuous weighting that naturally saturates between adjacent LoD levels, thereby preventing abrupt visibility changes. Please refer to the supplementary material for additional details of the LoD design.

This continuous opacity control allows alias-free and consistent rendering across scales. At each zoom step, new primitives are introduced to reconstruct appearance details, while existing layers remain frozen. Together, they form an adaptive generative hierarchy that maintains scene stability and progressively enhances geometric and semantic fidelity throughout the zoom-in process.

\subsection{Training Objective}
\begin{figure*}[t!]
  \centering
  \captionsetup[subfigure]{labelformat=empty, justification=centering}
  \scriptsize
  \vspace{-2.5em}
  \begin{subfigure}[t]{0.132\textwidth}
    \centering
    \includegraphics[width=\linewidth]{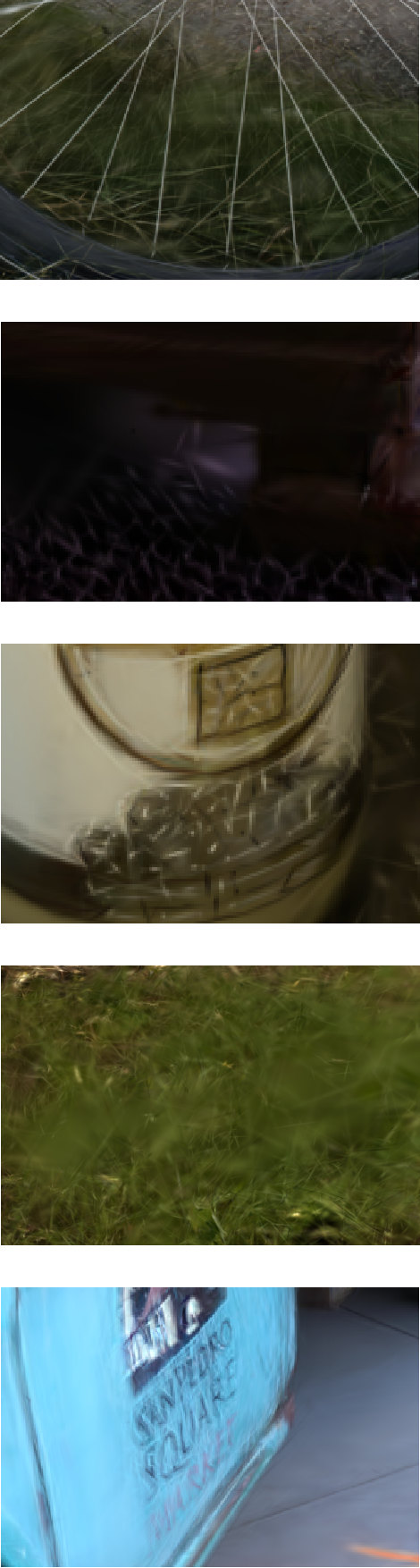}
    \caption{3DGS~\cite{kerbl20233d}}
  \end{subfigure}\hspace{1pt}
  \begin{subfigure}[t]{0.132\textwidth}
    \centering
    \includegraphics[width=\linewidth]{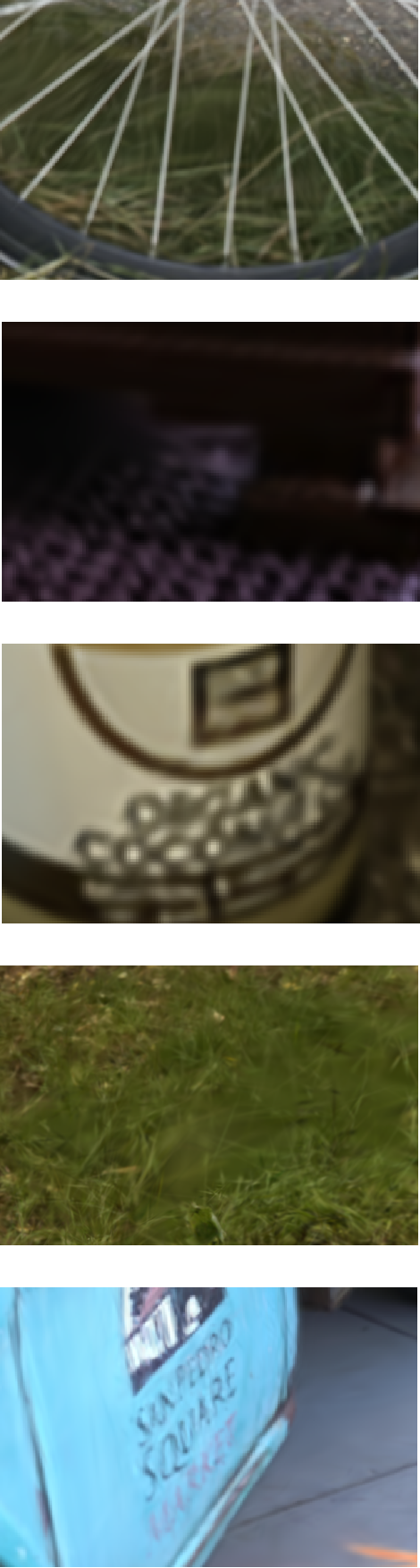}
    \caption{Mip-Splatting~\cite{yu2024mip}}
  \end{subfigure}\hspace{1pt}
  \begin{subfigure}[t]{0.132\textwidth}
    \centering
    \includegraphics[width=\linewidth]{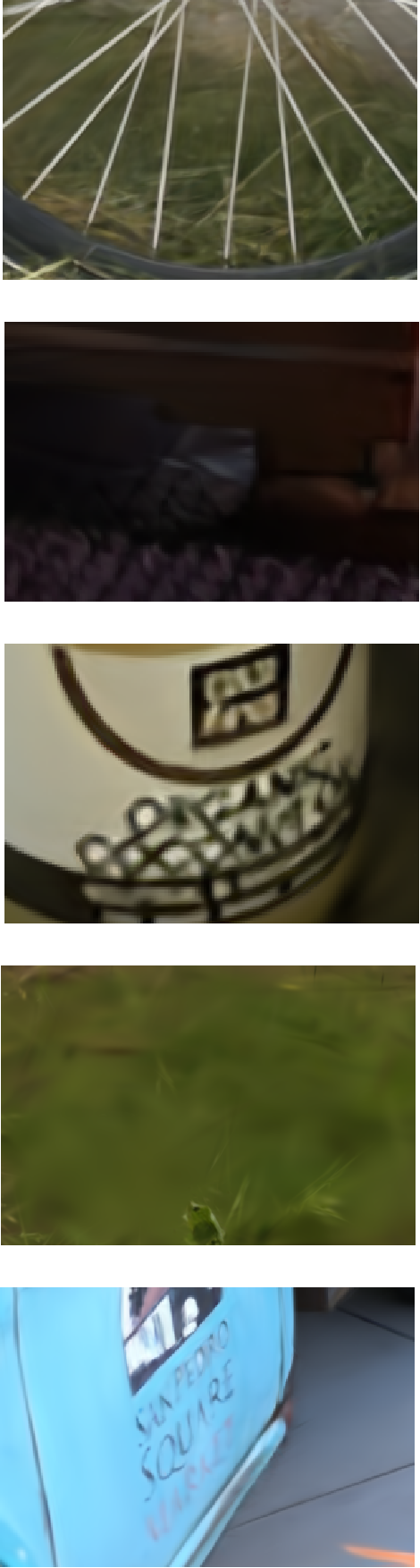}
    \caption{SuperGaussian~\cite{shen2024supergaussian}}
  \end{subfigure}\hspace{1pt}
  \begin{subfigure}[t]{0.132\textwidth}
    \centering
    \includegraphics[width=\linewidth]{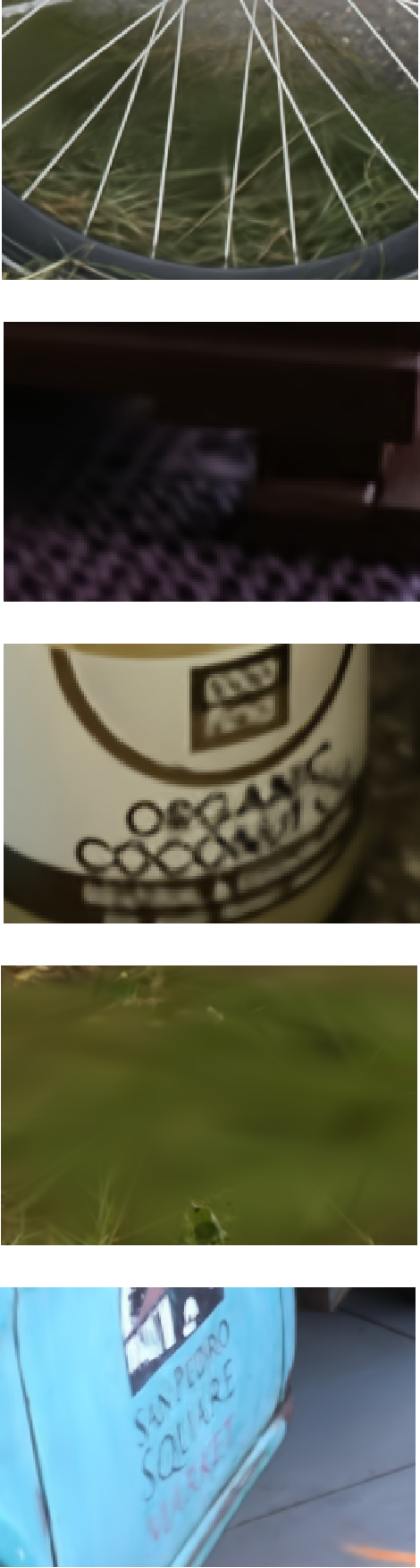}
    \caption{SRGS~\cite{feng2024srgs}}
  \end{subfigure}\hspace{1pt}
  \begin{subfigure}[t]{0.132\textwidth}
    \centering
    \includegraphics[width=\linewidth]{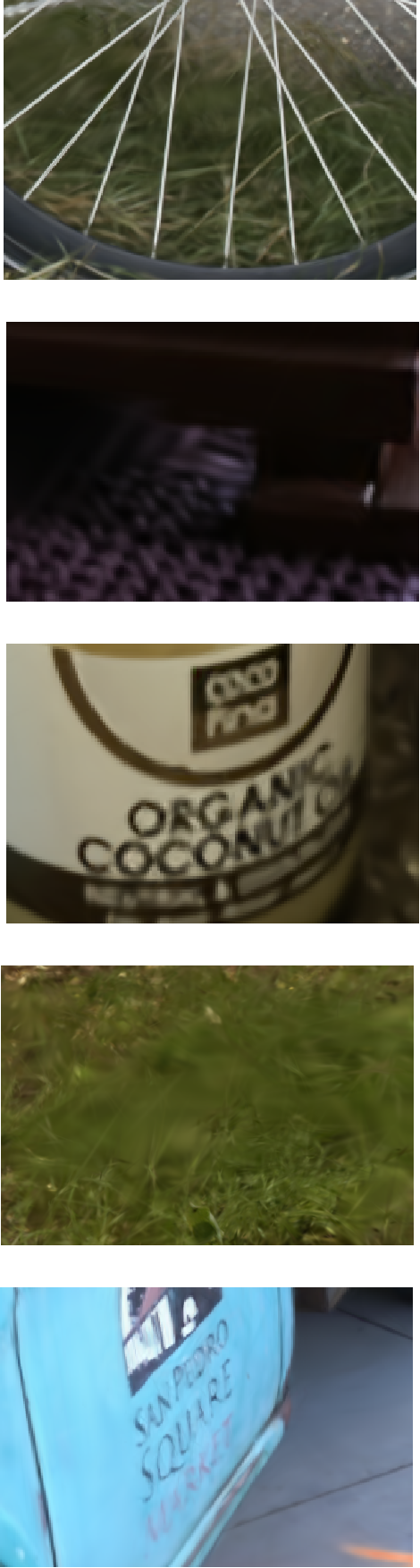}
    \caption{\hspace{-2.5mm}Sequence Matters~\cite{ko2025sequence}}
  \end{subfigure}\hspace{1pt}
  \begin{subfigure}[t]{0.132\textwidth}
    \centering
    \includegraphics[width=\linewidth]{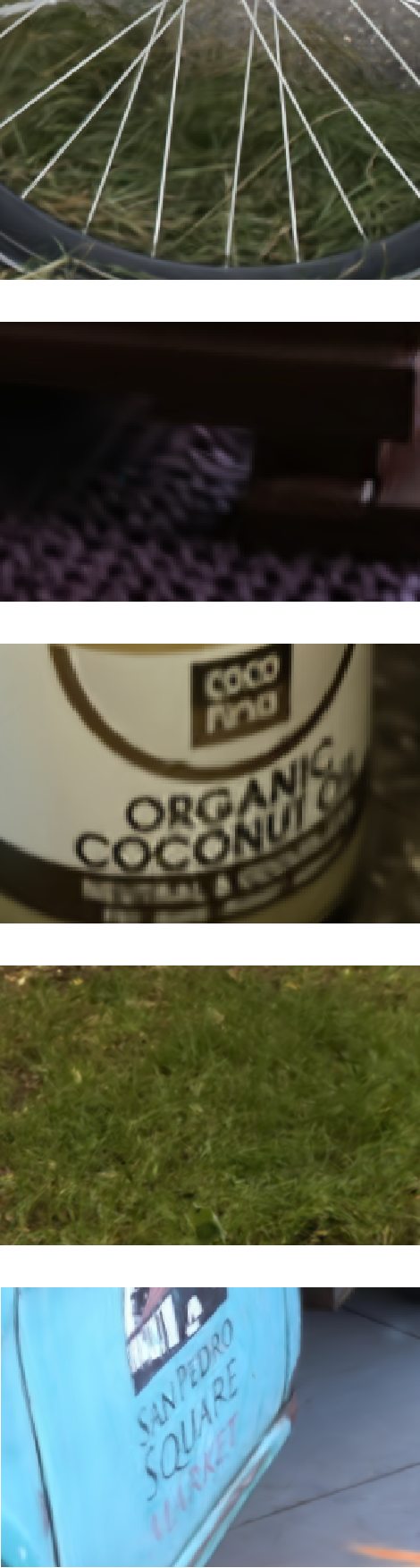}
    \caption{Ours}
  \end{subfigure}\hspace{1pt}
  \begin{subfigure}[t]{0.132\textwidth}
    \centering
    \includegraphics[width=\linewidth]{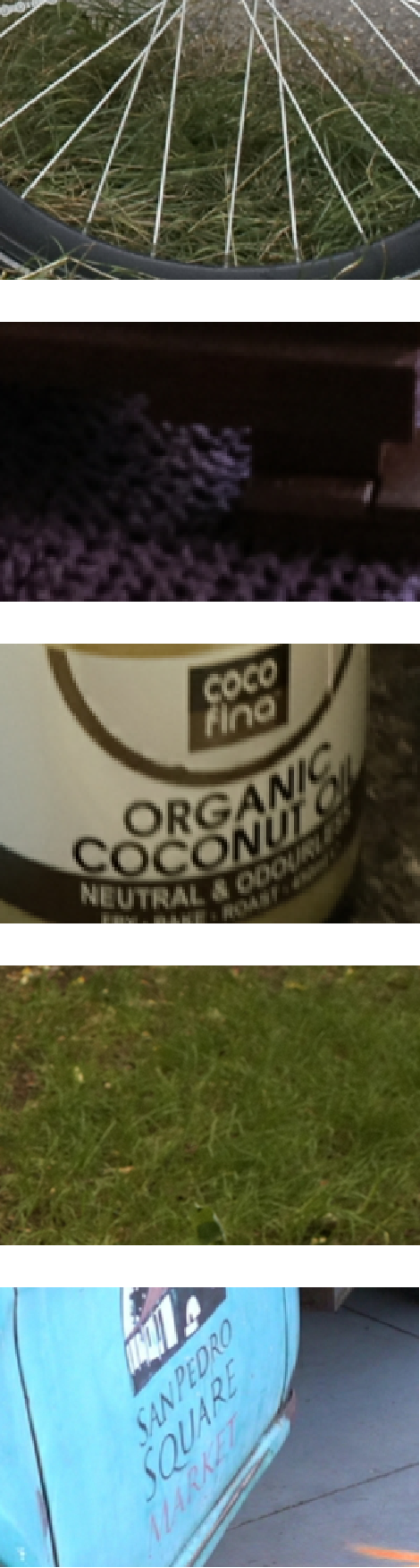}
    \caption{GT}
  \end{subfigure}
  \vspace{-0.8em}
  \caption{
    Qualitative comparison of $4\times$ super-resolution results. 
    Mip-Splatting reduces aliasing but lacks fine details; SuperGaussian, SRGS and Sequence Matters produces blurry textures; 
    Our method reconstructs sharper textures, cleaner edges, and more coherent structures across views, closely approaching the ground truth.
  }
  \label{fig:qualitative_x4}
\end{figure*}

\begin{table*}[t!]
  \vspace{-0.8em}
  \footnotesize
  \centering
  \setlength{\tabcolsep}{11pt}
  \begin{tabular}{@{}lcccccccc@{}}
    \toprule
    \multirow{2}{*}{\textbf{Method}}
      & \multicolumn{4}{c}{\textbf{Mip-NeRF360}}
      & \multicolumn{4}{c}{\textbf{Tanks\&Temples}} \\
    \cmidrule(lr){2-5} \cmidrule(lr){6-9}
      & PSNR$\uparrow$ & SSIM$\uparrow$ & LPIPS$\downarrow$ & FID$\downarrow$
      & PSNR$\uparrow$ & SSIM$\uparrow$ & LPIPS$\downarrow$ & \other{FID$\downarrow$} \\
    \midrule

    3DGS~\cite{kerbl20233d}
      & 20.64 & 0.634 & 0.385 & 60.48
      & 19.63 & 0.715 & 0.337 & \other{23.82} \\

    Mip-Splatting~\cite{yu2024mip}
      & 26.49 & 0.754 & 0.305 & \second{25.67}
      & 23.18 & 0.789 & 0.295 & \third{16.17} \\

    SuperGaussian~\cite{shen2024supergaussian}
      & 24.39 & 0.691 & 0.372 & 58.16
      & 21.40 & 0.752 & 0.343 & \other{39.62} \\

    SRGS~\cite{feng2024srgs}
      & \third{26.69} & \third{0.761} 
      & \third{0.301} & 33.97
      & \third{23.29} & \third{0.803}
      & \third{0.276} & \other{19.10} \\

    Sequence Matters~\cite{ko2025sequence}
      & \second{26.95} & \second{0.771}
      & \second{0.276} & \third{26.64}
      & \second{23.39} & \second{0.806}
      & \second{0.270} & \second{15.92} \\

    \textbf{Ours}
      & \best{\textbf{27.16}}
      & \best{\textbf{0.781}}
      & \best{\textbf{0.261}}
      & \best{\textbf{19.38}}
      & \best{\textbf{23.40}}
      & \best{\textbf{0.812}}
      & \best{\textbf{0.265}}
      & \best{\textbf{14.91}} \\
    \bottomrule
  \end{tabular}
  \vspace{-0.8em}
  \caption{Quantitative comparison on the Mip-NeRF360 ($1/8 \rightarrow 1/2$) and Tanks\&Temples ($1/4 \rightarrow 1$) datasets under the $4\times$ super-resolution setting. The best, second best and third best entries are marked in red, orange and yellow, respectively.}
  \label{tab:quantitative_x4}
  \vspace{-2em}
\end{table*}
Super-resolution inevitably introduces discrepancies between the synthesized high-resolution content and the structures observable in the low-resolution inputs. These inconsistencies can accumulate across zoom levels and destabilize the reconstruction. To alleviate this mismatch, we incorporate a subsampling-based dual-scale supervision that explicitly constrains the generated details to remain compatible with the underlying LR observations. Specifically, the rendered high-resolution image $R_i^{\text{hr}}$ is downsampled via bicubic interpolation as $R_i^{\text{lr}}$ which is then aligned with the corresponding low-resolution input $I_i^{\text{lr}}$. This enforces that the HR rendering does not deviate from the coarse-scale appearance when projected back to the LR domain.
\begin{equation}
\mathcal{L} = \lambda_\text{hr}\mathcal{L}_\text{rgb}(I_i^{\text{hr}}, R_i^{\text{hr}}) + \lambda_\text{lr}\mathcal{L}_\text{rgb}(I_i^{lr}, R_i^{\text{lr}})+\lambda_{\text{geo}}\mathcal{L_{\text{geo}}},
\end{equation}
where $\mathcal{L}_\text{rgb}$ is the RGB reconstruction loss combining $\mathcal{L}_1$ and with the D-SSIM term from 3DGS~\cite{kerbl20233d}, while $\mathcal{L_{\text{geo}}}$ is the geometry regularization loss from RaDe-GS~\cite{zhang2024rade}.

This dual-scale supervision effectively suppresses cross-scale conflicts introduced by super-resolution, ensuring that newly synthesized high-frequency details remain structurally consistent with the LR evidence throughout the progressive zoom-in process.



\section{Experiments}
\label{sec:experiment}

\subsection{Experiment Settings}
\noindent\textbf{Datasets.} 
We evaluate our method on two real-world benchmarks: Mip-NeRF360~\cite{barron2022mip} and Tanks\&Temples~\cite{knapitsch2017tanks}.
For the $4\times$ super-resolution task, the original Mip-NeRF360 images (approximately $3000{\times}4000$) are downsampled to $1/8$ resolution as low-resolution (LR) inputs and $1/2$ resolution as high-resolution (HR) targets.
For Tanks\&Temples, we use $1/4{\rightarrow}1$ resolution pairs.
Following standard practice, every eighth frame is reserved for testing while the remaining frames are used for training.
For the extreme zoom-in scenario with a magnification factor of $64$, we compute the intersection of camera frustums as region of interest and perform zoom-in generation within this region, which simplifies the setup without sacrificing generality. We then render smooth camera trajectories with focal lengths varying continuously from $1\times$ to $64\times$ to evaluate performance across large magnification ranges. 

\noindent\textbf{Metrics.}
For the $4\times$ super-resolution benchmark, we report standard full-reference metrics including PSNR~\cite{huynh2008scope}, SSIM~\cite{wang2004image}, LPIPS~\cite{zhang2018unreasonable}, and FID~\cite{heusel2017gans}.
FID measures the distributional distance between rendered and ground-truth images in the perceptual feature space.
For the extreme zoom-in scenario where no ground truth is available, we adopt no-reference perceptual quality metrics including CLIPIQA~\cite{wang2023exploring}, MUSIQ~\cite{ke2021musiq}, and NIQE~\cite{mittal2012making} to evaluate visual fidelity and realism.

\noindent\textbf{Baselines.}
We compare our approach against publicly released 3D super-resolution frameworks.
For 3DGS~\cite{kerbl20233d} and Mip-Splatting~\cite{yu2024mip}, models are trained directly on LR inputs and rendered at HR resolutions.
For SuperGaussian~\cite{shen2024supergaussian}, which applies video-based upsampling using VideoGigaGAN~\cite{xu2025videogigagan} along smooth camera trajectories, we replace VideoGigaGAN with BasicVSR~\cite{chan2021basicvsr} following the authors’ configuration, since the original VideoGigaGAN model has not been publicly released.
SRGS~\cite{feng2024srgs} employs a single-image SR backbone (SwinIR~\cite{liang2021swinir}), while Sequence Matters~\cite{ko2025sequence} adopts a video SR backbone (PSRT~\cite{shi2022rethinking}).
We follow their official implementations to generate SR-enhanced images and train corresponding 3DGS models on the refined datasets.
For the extreme zoom-in task, we compare only with SRGS~\cite{feng2024srgs} and Sequence Matters~\cite{ko2025sequence}, as the remaining baselines already exhibit substantial performance gaps at the $4\times$ setting.
\begin{figure*}[t!]
  \centering
  \scriptsize
  \vspace{-2.2em}
  \begin{subfigure}{\linewidth}
    \centering
    \begin{minipage}[t]{0.03\linewidth}
      \centering
      \vspace{15pt}
      \rotatebox{90}{View 1}\\[15pt]
      \rotatebox{90}{View 2}
    \end{minipage}%
    \begin{minipage}[t]{0.96\linewidth}
      \centering
      \makebox[0pt][c]{%
        \hspace{0.5em}Focal 1 \hspace{4em} Focal 2 \hspace{4em} Focal 3%
        \hspace{4.75em}
        Focal 1 \hspace{4em} Focal 2 \hspace{4em} Focal 3%
        \hspace{4.75em}
        Focal 1 \hspace{4em} Focal 2 \hspace{4em} Focal 3%
      }\\[4pt]
      \vspace{0.5pt}
      \includegraphics[width=\linewidth]{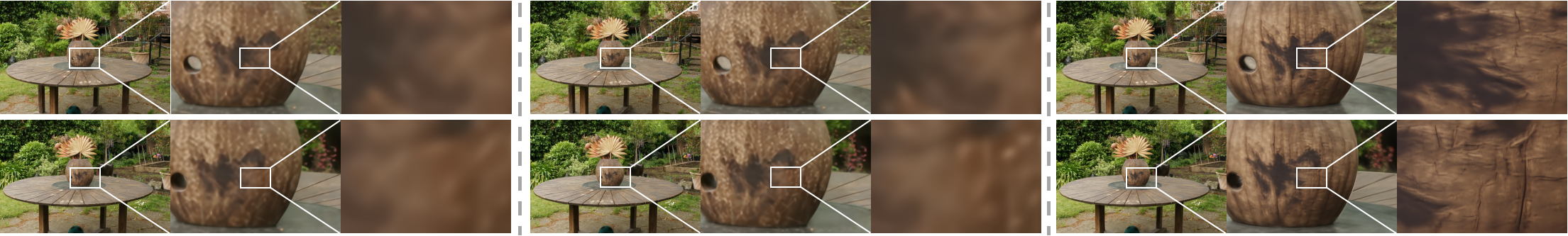}\\[2pt]
    \end{minipage}
    \vspace{-0.5em}
  \end{subfigure}

  \vspace{0.8em}

  \begin{subfigure}{\linewidth}
    \centering
    \begin{minipage}[t]{0.03\linewidth}
      \centering
      \vspace{10pt}
      \rotatebox{90}{View 1}\\[15pt]
      \rotatebox{90}{View 2}
    \end{minipage}%
    \begin{minipage}[t]{0.96\linewidth}
      \centering
      \vspace{2pt}
      \includegraphics[width=\linewidth]{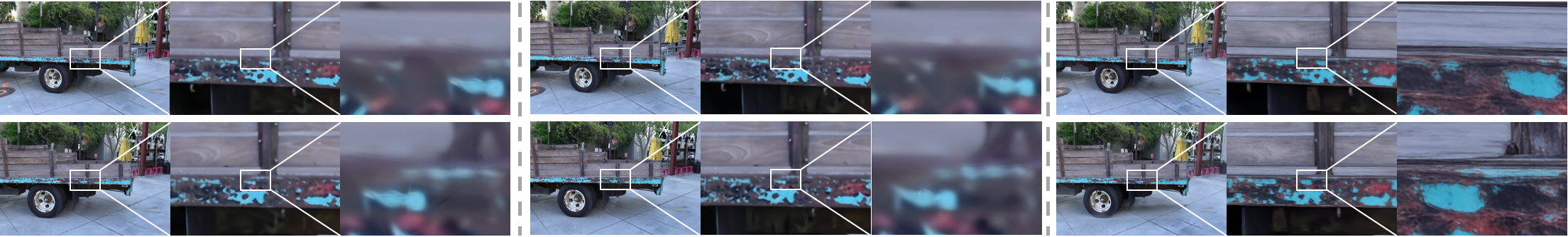}\\[2pt]
      \makebox[0pt][c]{%
        {\footnotesize SRGS \cite{feng2024srgs}}\hspace{16em}{\footnotesize Sequence Matters~\cite{ko2025sequence}}\hspace{16em}{\footnotesize Ours}%
      }
    \end{minipage}
    \vspace{-0.5em}
  \end{subfigure}

  \vspace{-0.8em}
  \caption{
    Qualitative comparison under extreme zoom-in across multiple focal levels and viewpoints. 
    Competing methods exhibit blurry, textureless results as zoom increases, 
    while our method preserves sharp, semantically consistent details and maintains geometric alignment across scales. 
  }
  \label{fig:qualitative_x64}
\end{figure*}
\begin{table*}[t!]
  \vspace{-0.8em}
  \centering
  \footnotesize
  \setlength{\tabcolsep}{11pt}
  \begin{tabular}{@{}lccc ccc ccc@{}}
    \toprule
    \multirow{2}{*}{\textbf{Method}} 
      & \multicolumn{3}{c}{CLIPIQA $\uparrow$} 
      & \multicolumn{3}{c}{MUSIQ $\uparrow$} 
      & \multicolumn{3}{c}{NIQE $\downarrow$} \\
    \cmidrule(lr){2-4} \cmidrule(lr){5-7} \cmidrule(lr){8-10}
      & $16\times$ & $32\times$ & $64\times$
      & $16\times$ & $32\times$ & $64\times$
      & $16\times$ & $32\times$ & \other{$64\times$} \\
    \midrule

    SRGS~\cite{feng2024srgs}
      & \second{0.272} & 0.246 & \second{0.346}
      & \second{32.49} & 21.57 & \second{17.27}
      & 8.35 & 11.74 & \other{15.54} \\

    Sequence Matters~\cite{ko2025sequence}
      & 0.267 & \second{0.247} & 0.302
      & 32.27 & \second{21.66} & 15.44
      & \second{7.82} & \second{11.50} & \second{15.25} \\

    \textbf{Ours}
      & \best{\textbf{0.347}}
      & \best{\textbf{0.382}} 
      & \best{\textbf{0.436}}
      & \best{\textbf{48.86}} 
      & \best{\textbf{38.12}} 
      & \best{\textbf{42.21}}
      & \best{\textbf{5.55}} 
      & \best{\textbf{6.43}} 
      & \best{\textbf{5.53}} \\
      
    \bottomrule
  \end{tabular}
  \vspace{-0.8em}
  \caption{Quantitative comparison under the extreme zoom-in setting (magnification factors of $16$, $32$, and $64$). 
  The super-resolution involved methods including SRGS~\cite{feng2024srgs} and Sequence Matters~\cite{ko2025sequence} are chosen for comparsion, while SuperGaussian \cite{shen2024supergaussian} fails to produce meaningful results under this extreme zoom-in regime with its default setting.
  The best and second best entries are marked in red and orange respectively. 
  }
  \label{tab:quantitative_extreme_zoom_in}
  \vspace{-2em}
\end{table*}

\noindent\textbf{Implementation Details.}
We train 3DGS with geometric regularization from RaDe-GS~\cite{zhang2024rade} for 30K iterations on LR inputs to obtain stable scene geometry.
DLoRAL~\cite{sun2025one} serves as our video SR backbone, in which the original flow-based warping is replaced by our depth-guided alignment.
For semantic detail synthesis, we employ Qwen-VL-2.5-3B-Instruct~\cite{bai2025qwen2} fine-tuned by Chain-of-Zoom~\cite{kim2025chain} as the vision-language prompt generator.
We set $\lambda_{hr}=0.6$, $\lambda_{lr}=0.4$, $\lambda_{geo}=0.05$, and use a per-step zoom factor of $s=4$.
All experiments are conducted on a single NVIDIA RTX 4090 GPU.

\noindent\textbf{Quantitative Results.}
As shown in Tab.~\ref{tab:quantitative_x4}, our method achieves the highest PSNR and SSIM and the lowest LPIPS and FID on both Mip-NeRF360 and Tanks\&Temples.

Compared with 3DGS~\cite{kerbl20233d} and Mip-Splatting~\cite{yu2024mip} which do not employ super-resolution, our approach can reconstruct richer fine-scale structures. In contrast to super-resolution–based baselines such as SuperGaussian~\cite{shen2024supergaussian}, SRGS~\cite{feng2024srgs}, and Sequence Matters~\cite{ko2025sequence}, our depth-guided feature warping achieves markedly stronger multi-view consistency. By mitigating the cross-view discrepancies often introduced by per-view or flow-based enhancement, it substantially reduces the conflicts accumulated during the 3D reconstruction process. The lower FID further reflects the improved stability and coherence of the reconstructed high-frequency details.

For the extreme zoom-in scenario (Tab.~\ref{tab:quantitative_extreme_zoom_in}), our method achieves the best performance across all no-reference metrics, including CLIPIQA, MUSIQ, and NIQE.
These results demonstrate the robustness of our framework in reconstructing semantically coherent details under large magnification, validating its ability to generalize beyond supervised resolution scales.

\noindent\textbf{Qualitative Results.}
As illustrated in Fig.~\ref{fig:qualitative_x4}, Mip-Splatting~\cite{yu2024mip} effectively suppresses aliasing artifacts compared with 3DGS~\cite{kerbl20233d}, yet its renderings still exhibit limited fine-scale structural and textural detail. SRGS~\cite{feng2024srgs}, which relies on a single-image super-resolution backbone, improves per-view sharpness but fails to maintain cross-view coherence, since each frame is enhanced independently without geometric alignment. SuperGaussian~\cite{shen2024supergaussian} and Sequence Matters~\cite{ko2025sequence} adopt flow-based feature warping for video super-resolution; however, inaccurate flow estimation often leads to inconsistent correspondence across views and degraded high-frequency details, particularly in regions with occlusion or large parallax. By contrast, our depth-guided alignment provides accurate multi-view correspondence, enabling the reconstruction of richer details, and more coherent structures across viewpoints.

Fig.~\ref{fig:qualitative_x64} further shows results under the extreme zoom-in setting across multiple focal levels and different viewpoints.
When the magnification factor becomes large, competing methods (SRGS~\cite{feng2024srgs}, Sequence Matters~\cite{ko2025sequence}) tend to produce over-smoothed textures and collapse fine semantic structures.
In contrast, our method preserves semantically consistent details even under large magnifications, producing natural materials, sharp edges, and coherent appearance across scales.

\subsection{Ablation Studies}
We conduct a series of ablation experiments to analyze the contribution of each component in our framework.

\noindent\textbf{Effectiveness of Depth-based Feature Warping.} We evaluate the temporal and cross-view consistency of super-resolved images using the Fréchet Video Distance (FVD)~\cite{unterthiner2018towards}. As reported in Tab.~\ref{tab:fvd_comparison}, our method achieves the lowest FVD scores on both Mip-NeRF360 and Tanks\&Temples, indicating superior temporal consistency.
When depth-guided feature warping is removed, FVD scores increase noticeably, demonstrating that geometry-based alignment effectively improves multi-view consistency compared to flow-based correspondence.

\noindent\textbf{Effectiveness of VLM Guidance.}
We further assess the role of the VLM-based semantic prompting. As shown in Fig.~\ref{fig:ablation_prompt}. 
Without prompt guidance, the reconstructed region exhibits semantic and material inconsistencies with the low-resolution inputs, producing mismatched textures or over-simplified surfaces. 
For example, the truck surface appears uniformly glossy rather than displaying the rust stains present in the input scene, indicating that the model enhances local contrast but fails to capture material semantics.
By contrast, incorporating VLM-inferred prompts provides scene-aware semantic priors that guide the SR model toward more plausible and coherent detail synthesis.
These observations highlight that semantic conditioning not only enriches perceptual realism but also helps maintain consistency with the global scene context.

\begin{table}[t]
  \footnotesize
  \centering
  \setlength{\tabcolsep}{6pt}
  \begin{tabular}{lcc}
    \toprule
    \textbf{Method}  & Mip-NeRF360 & Tanks\&Temples \\
    \midrule
    SuperGaussian~\cite{shen2024supergaussian} & 574.92 & 1941.06 \\
    SRGS~\cite{feng2024srgs}          & 175.89    & 205.82 \\
    Sequence Matters~\cite{ko2025sequence} & \second{165.74}    & \third{190.97} \\
    Ours w/o depth warping & \third{168.36} &  \second{180.45} \\
    Ours            & \best{\textbf{107.99}}     & \best{\textbf{79.98}} \\
    \bottomrule
  \end{tabular}
  \vspace{-0.5em}
  \caption{Fréchet Video Distance ($\downarrow$) of super-resolved images on Mip-NeRF360 and Tanks\&Temples datasets. The best, second best, and third best entries are marked in red, orange, and yellow, respectively.}
  \label{tab:fvd_comparison}
  \vspace{-1em}
\end{table}

\begin{figure}[t]
  \centering
  \setlength{\tabcolsep}{1pt} 
  \renewcommand{\arraystretch}{0.5}
  \begin{subfigure}[t]{0.22\textwidth}
    \centering
    \includegraphics[width=\linewidth]{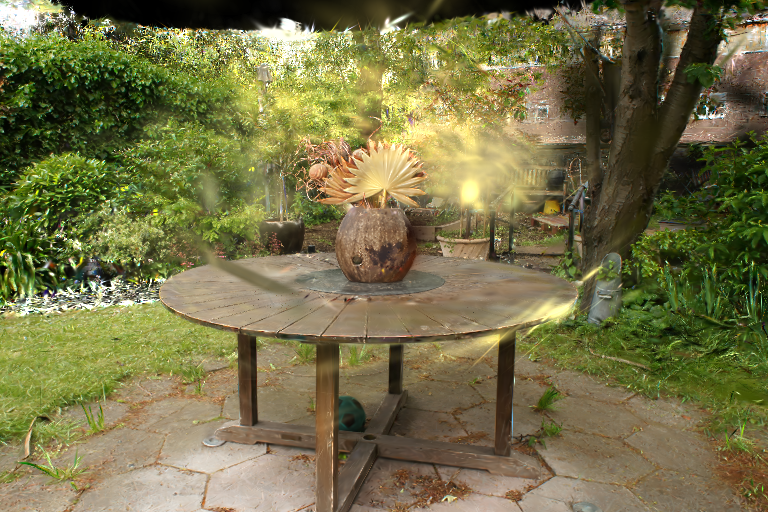}
    \caption{Zoomed out view without LoD}
  \end{subfigure}
  \hspace{8pt}
  \begin{subfigure}[t]{0.22\textwidth}
    \centering
    \includegraphics[width=\linewidth]{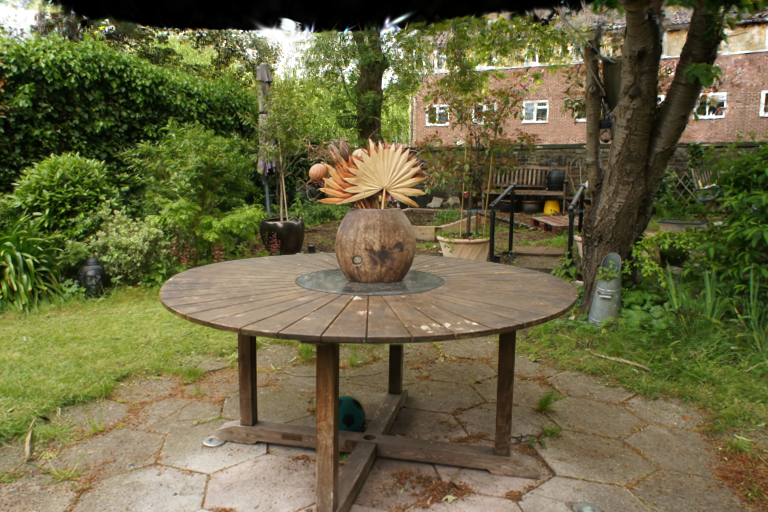}
    \caption{Zoomed out view with LoD}
  \end{subfigure}

  \begin{subfigure}[t]{0.22\textwidth}
    \centering
    \includegraphics[width=\linewidth]{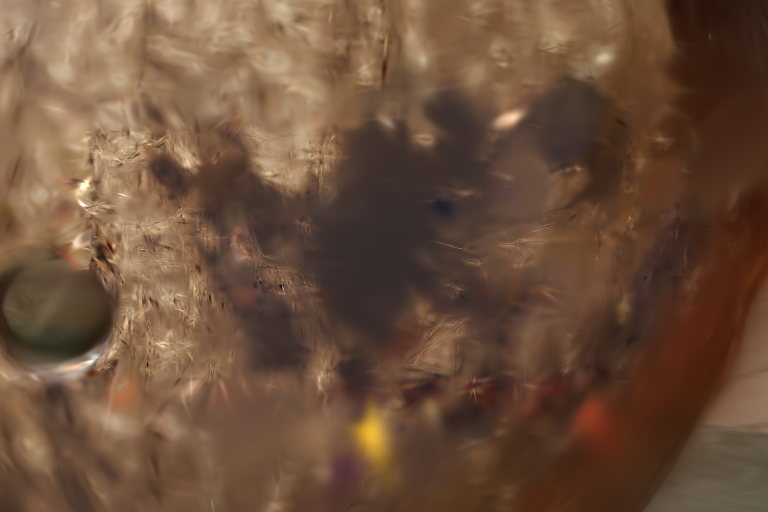}
    \caption{Zoomed in view without LoD}
  \end{subfigure}
  \hspace{8pt}
  \begin{subfigure}[t]{0.22\textwidth}
    \centering
    \includegraphics[width=\linewidth]{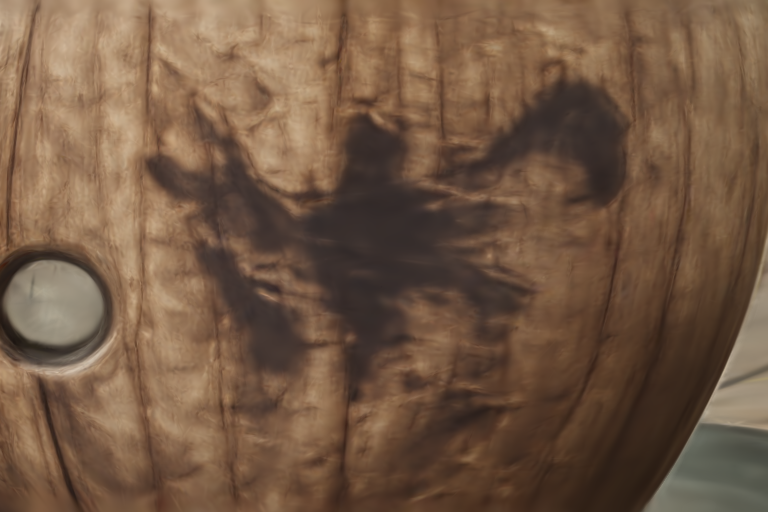}
    \caption{Zoomed in view with LoD}
  \end{subfigure}
  \vspace{-0.5em}
  \caption{
    Effectiveness of continuous LoD.
    Without LoD, optimizing a single Gaussian set across scales causes aliasing and semantic inconsistency.
  }
  \label{fig:ablation_lod}
  \vspace{-1em}
\end{figure}

\noindent\textbf{Effectiveness of Continuous Level-of-Detail.}
Fig.~\ref{fig:ablation_lod} compares renderings with and without the proposed continuous LOD hierarchy.
Because the super-resolved images at different scales inevitably exhibit slight inconsistencies, joint optimization under a shared representation leads to cross-scale conflicts and aliasing artifacts in both zoom-in and zoom-out renderings.
In contrast, our LoD hierarchy explicitly allocates separate Gaussian layers for different scales, allowing each level to specialize in its corresponding resolution.
This scale-aware organization and continuous adjustment effectively suppresses inter-scale interference and ensures smooth transitions across magnification levels.
\begin{figure}[t]
  \centering
  \setlength{\tabcolsep}{1pt} 
  \renewcommand{\arraystretch}{0.5}
  \begin{subfigure}[t]{0.22\textwidth}
    \centering
    \includegraphics[width=\linewidth]{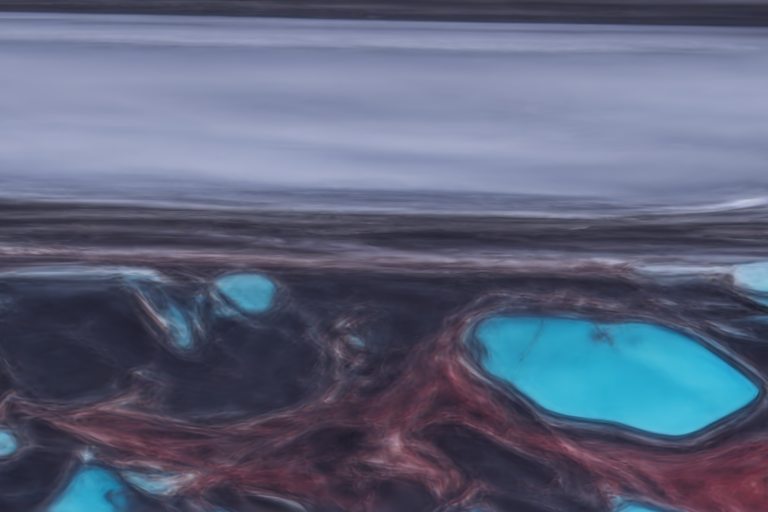}
    \caption{Without VLM prompt}
  \end{subfigure}
  \hspace{4pt}
  \begin{subfigure}[t]{0.22\textwidth}
    \centering
    \includegraphics[width=\linewidth]{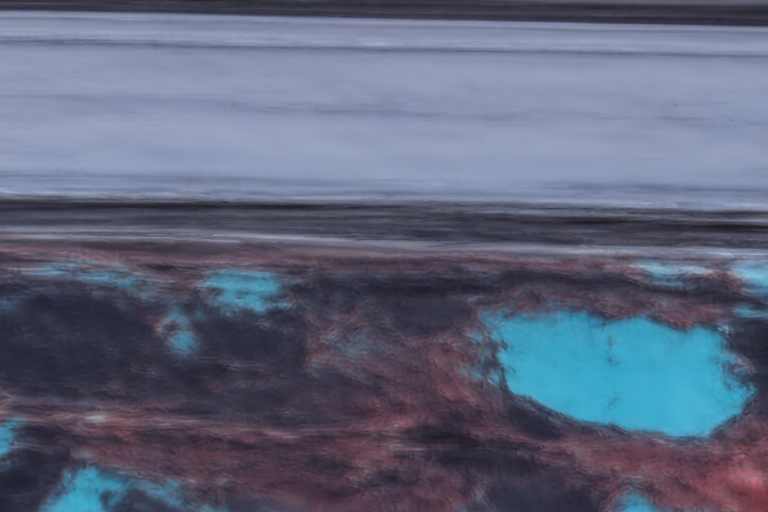}
    \caption{With VLM prompt}
  \end{subfigure}
  \vspace{-0.5em}
  \caption{
    Effectiveness of VLM guidance in detail synthsis.
    Without prompt guidance, the region becomes visually sharper but semantically inconsistent with the input (e.g. the truck surface loses its rusted texture).
  }
  \label{fig:ablation_prompt}
\end{figure}
\section{Conclusion}
\label{sec:conclusion}
We have presented GaussianZoom, a generative zoom-in 3D reconstruction framework that integrates geometry-consistent scene modeling with semantic detail refinement. A multi-view consistent super-resolution module, built upon depth-based feature warping and VLM-driven detail synthesis, ensures accurate cross-view correspondence and fine-scale appearance enhancement, while an expandable continuous LoD hierarchy dynamically modulates Gaussian visibility to enable smooth, alias-free rendering across scales. Experiments on various datasets show that GaussianZoom achieves superior perceptual quality and multi-view consistency under extreme zoom-in.

\noindent\textbf{Limitations.}
Despite its strengths, our method encounters difficulties at very high magnifications (e.g., $\times1024$), where current vision-language models struggle to infer coherent structures, leading to semantically weak textures. Future work will investigate more capable content creative zoom-in approaches to enable seamless transitions from cosmic-scale environments down to microscopic and molecular scenes.

\section{Acknowledgements}
\label{sec:acknowledge}
\vspace{-0.5 em}
We thank Boming Zhao for helpful discussions.
This work was supported by the National Key R\&D Program of China (2024YFC3811000), the NSFC (No.~62572425 and No.~624B2132), Information Technology Center, and State Key Lab of CAD\&CG, Zhejiang University.

\clearpage
{
    \small
    \bibliographystyle{ieeenat_fullname}
    \bibliography{main}
}


\end{document}